\documentclass[pdflatex,sn-aps]{sn-jnl}

\usepackage{graphicx}%
\usepackage{multirow}%
\usepackage{amsmath,amssymb,amsfonts}%
\usepackage{amsthm}%
\usepackage{mathrsfs}%
\usepackage[title]{appendix}%
\usepackage{xcolor}%
\usepackage{textcomp}%
\usepackage{manyfoot}%
\usepackage{booktabs}%
\usepackage{algorithm}%
\usepackage{algorithmicx}%
\usepackage{algpseudocode}%
\usepackage{listings}%
\usepackage{float}

\theoremstyle{thmstyleone}%
\theoremstyle{thmstyletwo}%

\theoremstyle{thmstylethree}%

\begin{document}

\title[Article Title]{VisionCLIP: An Med-AIGC based Ethical Language-Image Foundation Model for Generalizable Retina Image Analysis}


\author[1]{\fnm{Hao} \sur{Wei}}\email{haowei@link.cuhk.edu.hk}
\author[1]{\fnm{Bowen} \sur{Liu}}
\author[1]{\fnm{Minqing} \sur{Zhang}}
\author[1]{\fnm{Peilun} \sur{Shi}}


\author*[1]{\fnm{Wu} \sur{Yuan}}\email{wyuan@cuhk.edu.hk}

\affil*[1]{\orgdiv{Department of Biomedical Engineering}, 

\orgname{The Chinese University of Hong Kong},  \city{Hong Kong SAR}, \country{China}}

\abstract{Generalist foundation model has ushered in newfound capabilities in medical domain. However, the contradiction between the growing demand for high-quality annotated data with patient privacy continues to intensify. The utilization of medical artificial intelligence generated content (Med-AIGC) as an inexhaustible resource repository arises as a potential solution to address the aforementioned challenge. Here we harness 1 million open-source synthetic fundus images paired with natural language descriptions, to curate an ethical language-image foundation model for retina image analysis named VisionCLIP.  VisionCLIP achieves competitive performance on three external datasets compared with the existing method pre-trained on real-world data in a zero-shot fashion. The employment of artificially synthetic images alongside corresponding textual data for training enables the medical foundation model to successfully assimilate knowledge of disease symptomatology, thereby circumventing potential breaches of patient confidentiality.}




\maketitle

\section{Introduction}\label{sec1}

\section*{Introduction}
Recent advances in medical foundation AI models exhibit an enhanced capacity to readily accommodate a broader range of tasks, leveraging their generalist intelligence rather than being confined to narrow objectives~\cite{moor2023foundation}. Among them, ophthalmic foundation models have been demonstrated for a wide range of clinical applications, including disease diagnosis, screening, and forecasting, providing accuracy and precision comparable to or beyond ophthalmologists~\cite{qiu2023visionfm,zhou2023foundation,dai2023deepdr,shi2023generalist}. However, all the previous large AI models for ophthalmology were developed based on hundreds of thousands or millions of clinical data from screen projects, outpatient and inpatient eye centres. The presence of heterogeneous data sources poses a significant challenge for organizations and data sharing. Additionally, the development of data-driven medical foundation models raises serious concerns regarding the protection of patients' private information. There is substantial evidence indicating that large AI models can accurately infer a patient's age, gender~\cite{poplin2018googleprediction}, and even race~\cite{gichoya2022airace} from medical image inputs. This poses a risk to privacy and goes against the original purpose of developing large medical AI models~\cite{seyyed2021underdiagnosis}. Generative Artificial Intelligence is a potential solution to address these issues by achieving an idealized distribution of medical data through random noise sampling. This approach enables access to unrestricted data in terms of both privacy and volume~\cite{ning2023generative,shilo2020axes}.

Therefore, we try to explore the utilization of generated data in this study and to figure out whether models trained with fully synthetic data can achieve comparable performance in downstream tasks. Specifically, we first involve the SynFundus-1M~\cite{shang2023synfundus} dataset, which contains 1 million synthetic retinal fundus images and the corresponding diagnostic descriptions (Chinese), to train a language-image foundation model. The synthesis process that starts from the pure noise image ensures the avoidance of any ethical and privacy issues in images, such as age, gender, and race. Then, to supervise the image representation learning, we adopted the Contrastive Language-Image Pre-Training (CLIP)~\cite{radford2021clip} recipe to leverage the rich semantics in corresponding textual descriptions. During experiments, comprehensive evaluations on both public and private datasets are conducted to evaluate the ability of the trained image encoder in zero-shot learning scenarios.

To summarize, our main contributions are the following: (1) We proposed the first ethical-free foundation model for retinal image analysis; (2) Through extensive experiments, we validated the synthetic data also can achieve comparable performance compared with models trained on clinical data.

\begin{figure}
    \centering
    \includegraphics[width=\textwidth]{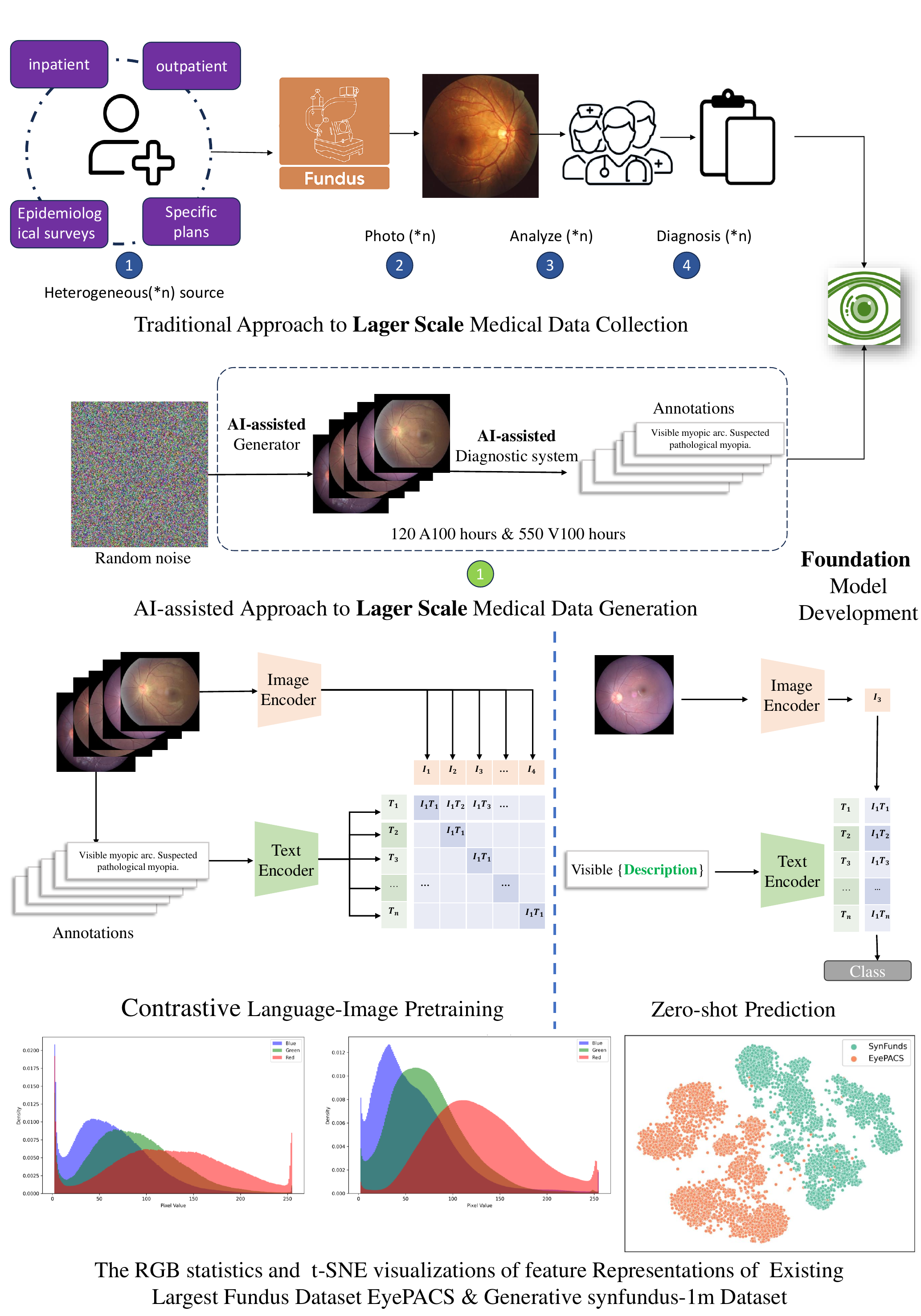}
    \caption{Overview of the pipeline involved in our study.}
    \label{fig_overview}
\end{figure}





\section{Methods}\label{sec2}

\subsection{Datasets}
In our study, we employ the SynFundus-1M\cite{shang2023synfundus} dataset, an expansive and meticulously annotated collection consisting of over one million synthetic fundus images. This dataset spans eight types of retinal diseases and is characterized by an innovative annotation strategy that includes four readability labels for key image regions, and the corresponding textural descriptions. Engineered with the aid of a Denoising Diffusion Probabilistic Model named SynFundus-Generator, which itself was trained on a comprehensive set of 1.3 million authentic fundus images from varied clinical settings, SynFundus-1M stands as a testament to the potential of synthetic data in overcoming conventional challenges of data privacy and annotation costs in medical image research. During experiments, we randomly split the whole dataset into three parts with 8:1:1 ratios as training, validation and test set.

\subsection{Training pipeline}
Annotation in Synfundus-1M was provided in Chinese, and we leveraged Contrastive Vision-Language Pretraining in Chinese version for foundation model training~\cite{yang2022chineseclip}, where the whole pipeline is shown in Fig.\ref{fig_overview}. Specifically, the image encoder utilized is ResNet50, while the text encoder is based on RoBERTa-wwm-ext-base-chinese. Both the image and text encoders were initialized randomly during the training of VisionCLIP. In the VisionCLIP training phase, the image and text embedding vectors, corresponding to each other, were extracted by the two encoders. All paired vectors were then optimized to minimize their distance using a contrastive approach.

\section{Experiments}
In this section, we mainly evaluate the zero-shot performance of our trained encoder on the regular diagnosis task: the diabetic grading and glaucoma screening.

\subsection{Datasets}
In our experiments, we utilized three pivotal retinal fundus image datasets, each offering unique insights into ocular conditions and aiding in the development of automated diagnostic tools. Below, we briefly introduce these datasets:
\begin{itemize}
    \item MESSIDOR\cite{decenciere2014feedback}: The Messidor dataset is a well-known collection of 1200 retinal images, established to facilitate research in computer-assisted diagnosis of diabetic retinopathy. Images were captured with a 45-degree field of view and vary in resolution, providing a comprehensive resource for evaluating segmentation algorithms and indexing techniques related to retinal ophthalmology2.
    \item FIVES\cite{jin2022fives}: FIVES (Fundus Image Dataset for Artificial Intelligence-based Vessel Segmentation) comprises 800 high-resolution, multi-disease color fundus photographs, each meticulously annotated at the pixel level and also the image level. 
    \item REFUGE\cite{orlando2020refuge}: The REFUGE (Retinal Fundus Glaucoma Challenge) dataset is currently the largest of its kind, containing 1200 fundus images with ground truth segmentations and clinical glaucoma labels. It serves as a unified framework for evaluating automated methods for glaucoma assessment from fundus photographs, providing a rich source of data for optic disc and cup segmentation tasks.
\end{itemize}

\subsection{Related works}
In our experiment, we involved three methods in the mentioned datasets to evaluate their zero-shot performance. Specifically, Radford et al.\cite{radford2021clip} introduced a paradigm where visual models are trained via natural language (CLIP), facilitating a flexible and transferable learning process. Zhang et al.\cite{zhang2023biomedclip} advanced this concept with BiomedCLIP, a model pre-trained on a vast dataset of scientific image-text pairs, specifically tailored for biomedical applications. Lastly, Silva-Rodriguez et al.\cite{silva2023flair} presented FLAIR, a model that integrates expert knowledge through text supervision to enhance the interpretation of retinal images.

\subsection{Results}
The Table. \ref{tab_1} demonstrates the results of all involved methods on three datasets. From this table, our VisionCLIP possesses the capability to process a wide array of retinal images, transcending categorical limitations, without the necessity for additional explicit training. Another foundational language-image model, FLAIR~\cite{silva2023flair} has been selected as a baseline. This serves to illustrate the efficacy of the proposed synthetic-based foundational model. All experimental datasets and evaluation metrics are set based on previous models to ensure a fair comparison.




\begin{table}[]
\label{tab_1}
\caption{Zero-shot performance (accuracy) on the three external datasets.}
\begin{tabular}{@{}cccc@{}}
\toprule
Method     & \multicolumn{3}{c}{Dataset}                      \\ \cmidrule(l){2-4} 
           & MESSIDOR       & FIVES          & REFUGE         \\ \midrule
CLIP\cite{radford2021clip}       & 0.237          & 0.250          & 0.470          \\
BiomedCLIP\cite{zhang2023biomedclip} & 0.224          & 0.416          & 0.540          \\
FLAIR\cite{silva2023flair}      & 0.545          & 0.732          & 0.899          \\
FLAIR\_EK\cite{silva2023flair}  & \textbf{0.604} & 0.735          & 0.920          \\
VisionCLIP      & 0.431          & \textbf{0.739} & \textbf{0.925} \\ \bottomrule
\end{tabular}
\end{table}

\section{Conclusion}
In conclusion, the development of VisionCLIP represents a significant advancement in the field of medical artificial intelligence for retina image analysis. By harnessing 1 million synthetic fundus images paired with natural language descriptions, VisionCLIP has successfully addressed the challenges of data privacy and annotation costs in medical image research. This foundation model not only achieves competitive performance on external datasets but also ensures patient confidentiality by avoiding the use of real-world data that may compromise privacy.

Through extensive experiments and evaluations, VisionCLIP has demonstrated its ability to process a wide source of retinal images without the need for additional explicit training, showcasing its generalizability and efficiency in automated diagnostic tasks. The model's utilization of artificially synthetic images alongside textual data has enabled it to assimilate knowledge of disease symptomatology effectively, paving the way for improved accuracy and precision in retina image analysis.

Overall, VisionCLIP stands as a testament to the potential of synthetic data in overcoming traditional challenges in medical image analysis. Its ethical approach to data utilization, coupled with its high performance and patient privacy protection, positions it as a valuable tool for enhancing the capabilities of medical foundation AI models in real-world clinical settings.

\bibliography{sn-bibliography}

\end{document}